\documentclass[10pt,twocolumn,letterpaper]{article}

\usepackage{cvpr}              % To produce the CAMERA-READY version
\usepackage{tabularx}
\graphicspath{ {./images/} }
\usepackage[dvipsnames]{xcolor}

\definecolor{cvprblue}{rgb}{0.21,0.49,0.74}
\usepackage[pagebackref,breaklinks,colorlinks,citecolor=cvprblue]{hyperref}

\def\paperID{8} % *** Enter the Paper ID here
\def\confName{CVPR}
\def\confYear{2024}

\title{Segment Anything Model for Grain Characterization in Hard Drive Design}

\author{Kai Nichols$^{1}$ \quad Matthew Hauwiller$^{1}$ \quad Nicholas Propes$^{1}$ \quad Shaowei Wu$^{1,2}$\\ \quad Stephanie Hernandez$^1$ \quad Mike Kautzky$^{1}$ \vspace{0.2em} \\
{$^1$Seagate Technology} \quad
{$^2$Department of Physics, University of Minnesota} \quad \\
}

\begin{document}

\maketitle
\begin{abstract}
Development of new materials in hard drive designs requires characterization of nanoscale materials through grain segmentation. The high-throughput quickly changing research environment makes zero-shot generalization an incredibly desirable feature. For this reason, we explore the application of Meta's Segment Anything Model (SAM) to this problem. We first analyze the out-of-the-box use of SAM. Then we discuss opportunities and strategies for improvement under the assumption of minimal labeled data availability. Out-of-the-box SAM shows promising accuracy at property distribution extraction. We are able to identify four potential areas for improvement and show preliminary gains in two of the four areas.

\end{abstract}    
\section{Introduction}
\label{sec:intro}

Data is being generated and utilized at an ever-increasing pace, and this demand for data necessitates cheap, mass capacity data storage. Data storage has not been able to keep up with the pace of data creation, and by 2025, world-wide storage capacity will only be able to store 10\% of the digital data created \cite{Reinsel2017}. New Hard Disk Drive (HDD) advances such Heat Assisted Magnetic Recording (HAMR) can increase the density of information on each disk, promising capacity increases of 2-3 times current HDD offerings \cite{Kautzky2018}. HAMR requires nanoscale characterization and control of materials to generate structures with the right magnetic, plasmonic, photonic, and electronic properties.

The grain structure is the same length scale as the nanodevices and can impact the material's properties \cite{Dillon2016}, so high-throughput techniques with nanoscale grain sensitivity are needed for both research and production. High-resolution grain characterization tools such as Atom Probe Tomography and Scanning Transmission Electron Microscopy lack the throughput for high volume data collection and entail destructive FIB cuts of the head \cite{Herbig2015}. Bulk techniques like X-Ray Diffraction average across a sample and miss spatial information. Scanning Electron Microscopy has an optimal balance of spatial resolution and throughput for grain characterization \cite{McIntosh1999}, but there remain grain boundary contrast challenges due to the vertically aligned grains.

Quantifying the grain structure requires segmentation of the grains, but rule-based segmentation models fail due to weak grain boundary contrast and contrast fluctuations across the image. Previous work has gotten around this problem by training neural networks on hundreds of hand-labeled grains \cite{Patrick2023,Shi2022}. This method works well for segmenting grain images if the material and image conditions are not changing, but in the fast-paced semiconductor research environment, processing conditions and material composition may be changing on even a wafer-to-wafer basis. A robust grain segmentation model is needed to avoid re-labelling and re-training for each new change in the process.

For this reason, we chose to investigate the use of Meta's Segment Anything Model (SAM), a large pre-trained general segmentation model \cite{Kirillov2023}. SAM has been shown to have some zero-shot segmentation performance in microscopy images \cite{Larsen2023,Chauveau2023}. This gives us the flexibility of a neural network across different imaging conditions while minimizing the need for hand-labeled images. The characteristic of Segment Anything as a prompt-based, ambiguity-aware instance segmentation model allows us to optimize our final output based on domain-specific patterns. We investigate the performance of SAM at grain property extraction and discuss paths for improvements in prompt point selection, prompt engineering, post-processing, and weight fine-tuning.
\section{Evaluating SAM for Grain Characterization}
\label{sec:segmentation}

\begin{figure*}
  \centering
  % \begin{subfigure}{0.44\linewidth}
  %     \includegraphics[width=\linewidth]{segmentation.png}
  % \end{subfigure}
  % \hfill
  % \begin{subfigure}{0.50\linewidth}
  %     \includegraphics[width=\linewidth]{property.png}
  % \end{subfigure}
  \includegraphics[width=0.89\linewidth]{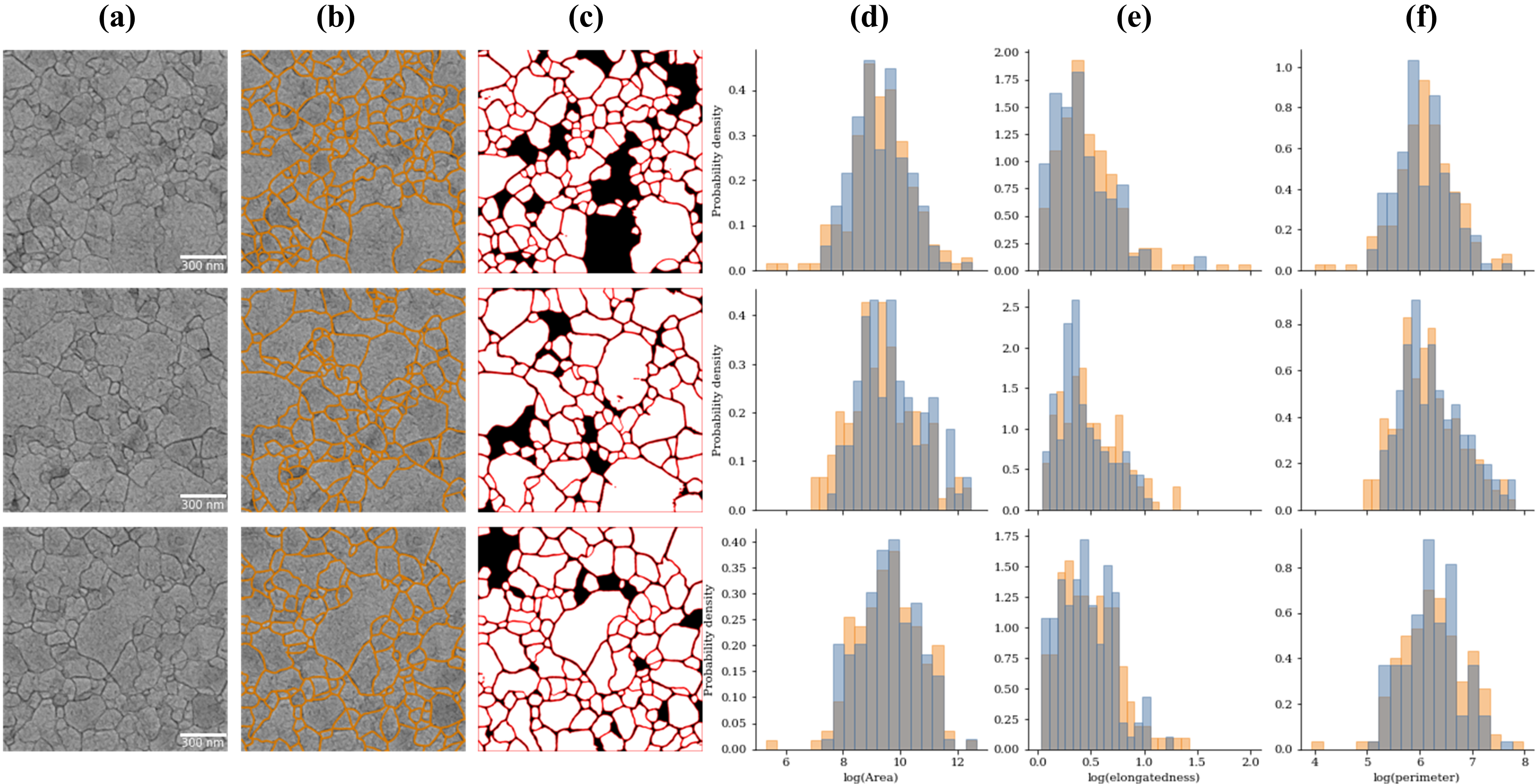}
  \caption{Segmentation with Automatic Mask Generator. (a) SEM images. (b) Hand drawn tracings. (c) SAM AMG + mask-level NMS segmentation. Masks are white with red outline. Where no masks were found is black. (d-f) Density Histograms of microstructure properties for hand-labeled grains (orange) and predicted grains (blue).}
  \label{fig:segmentation}
  \end{figure*}

In this section, we introduce our dataset. We then evaluate the performance of Meta's Automatic Mask Generator (AMG) pipeline at extraction of key grain microstructure properties \cite{Kirillov2023}. This pipeline uses the Segment Anything Model for whole-image segmentation. We then break down potential sources of error that could impact the pipeline's segmentation performance. 

\subsection{Data}
A dataset of 5 SEM images from one gold film wafer was used for this analysis. The gold film was deposited by Physical Vapor Deposition in Ultra-High Vacuum and then imaged on a KLA eDR7380 Scanning Electron Microscope. Backscattered electron imaging mode was used to collect the images.

The labels were obtained by a single annotator hand tracing over the images using a drawing tablet. Multiple annotators tend to disagree, so hand-drawn labels are not a perfect truth. The hand-labeling results in 783 grains.

\subsection{Segmentation Method}
\label{sec:segmentation_method}
SAM takes a prompt and image and outputs a boolean segmentation mask. The prompt can be any combination of a set of foreground and background points, a box, and a low resolution mask. It can return one output or three. The multi-mask output mode allows it to handle ambiguity in the output of simple prompts. The most common prompt type is a single foreground prompt point used to identify the object at that location on the image.

The Automatic Mask Generator (AMG) is Meta's method for full-image fully automatic instance segmentation using SAM. It consists of three main steps. First, is the extraction of segmentation masks over multiple overlapping zoomed-in crops using a grid of single foreground points as the prompt input. The zoom-in crops strategy is done to improve mask quality for smaller masks. The second step is mask post-processing to remove known common errors, namely small spurious components and small spurious holes. Finally, is the use of mask quality metrics namely predicted intersection over union (IoU) and stability. IoU is a commonly used metric for segmentation tasks. Stability is the sensitivity of the output to changes in the cutoff value used to binarize the prediction. These quality metrics are used to filter out bad masks and resolve duplicates using box-level non-maximum supression (NMS). This code is provided at their repo \href{https://github.com/facebookresearch/segment-anything}{https://github.com/facebookresearch/segment-anything.} 

We use the default settings of AMG with the pre-trained model weights for the ViT-h \cite{dosovitskiy2021an} architecture. The resulting masks are then filtered with mask-level NMS with SAM's predicted IoU scores as the criteria. This is a stricter constraint on overlapping objects than the AMG provides, but is necessary to a guarantee maximum overlap threshold consistent with grain boundary expectations. The inputs and resulting output can be seen in \Cref{fig:segmentation}a-c. Holes in the final output happen for two reasons. Some small grains are not detected, and the model asymmetrically struggles with some low contrast boundaries. This means that boundaries are only detected from foreground point prompts on one side of the boundary. Other miss-segmentation is due to low-contrast boundaries being missed or intra-grain texture being recognized as a boundary.

\begin{figure*}[t]
  \centering
  \includegraphics[width=0.89\linewidth]{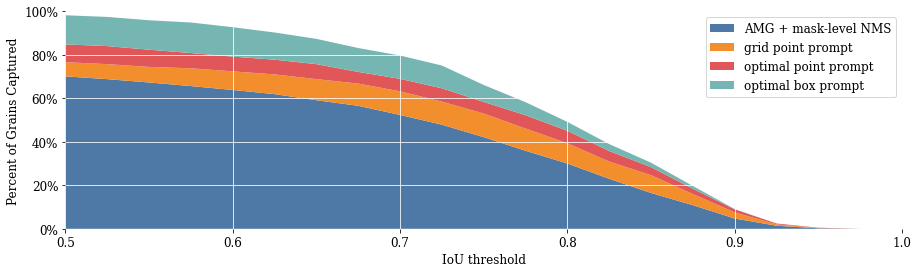}
  \caption{Comparison of ability of SAM to capture hand-labeled grains under increasingly generous conditions.}
  \label{fig:cause}
\end{figure*}

\subsection{Property Extraction}
To evaluate effectiveness of the AMG at property extraction we compare the distributions of area, elongatedness, and perimeter between hand-labeled and model results. The results are shown in \Cref{fig:segmentation}d-f. There is good alignment between the two methods. However, there are some consistent biases in the prediction across all images. The lower tail of area, lower tail of perimeter, and upper tail of elongatedness are shorter as these grains are missing.

\subsection{SAM's Limitations}
\label{sec:limitations}
We consider four reasons for the inability of the segmentation method to capture every grain and outline a method for quantifying the number of grains not captured attributable to each reason. First, a good mask for the grain may have been found but filtered out by the post-processing steps. We save these masks before post-processing. Second, the grain may not have contained any of the prompt points. We consider the masks of 50 random prompt points per grain. Third, we consider grains that could be captured with a better prompt input. These are extracted via a box prompt aligned with the bounding box of the hand-labeled grain. They represent an approximate upper limit on the performance of the model via prompting without changing any model weights. Finally, there are those can not be directly captured by the default SA model. In \Cref{fig:cause}, we compare the percent of grains falling into each of these categories at different IoU thresholds. At a 0.7 IoU threshold, there are 27.34\% of grains which could potentially be recovered without changing SAM's weights.
% \section{Maximizing Performance with Pre-Trained SAM} % alternatively: training-free improvements
\section{Training-Free Improvements}
\label{sec:pre_trained}
The AMG pipeline allows for three areas to optimize without adjusting model weights. One can change the input image, the input prompt(s), and how output masks are post-processed. We believe that incorporating domain specific knowledge into these model pipeline choices can improve our results. There are many inherent characteristics of the grain structure that can be leveraged in making these choices. We have listed some in \Cref{tab:characteristics}. In this section, we present our initial investigation into improvements via input prompt point selection and output mask overlap resolution. We will also discuss other potential ideas for improvement.

\subsection{Prompt-Selection}

\begin{table}
  \centering
  \begin{tabular}{m{4.5cm} m{2.5cm}}
    \toprule
    Method & mIoU \\
    \midrule
     18x18 grid & 0.4901 \\
     $\sim$ 300 iterative points per image  & 0.5609  \\
     \midrule
     difference & + 0.0709\\
     \hspace{1em} Bootstrapped 95\% CI & (0.0524, 0.0900)\\
    \bottomrule
  \end{tabular}
  \caption{Comparison of prompt point selection methods}
  \label{tab:prompt}
  \vspace{-0.5cm}
\end{table}

Our analysis in \Cref{sec:limitations} shows 2-8\% of grains can be found if an optimal foreground prompt point is selected. Due to the expected long-tailed distribution of grain size, grid prompt point selection is ill-suited to guaranteeing each grain will have at least one prompt point within its boundaries. We test a process of iteratively selecting prompt points where holes were left in the previous step.

In \Cref{tab:prompt}, we compare iterative and grid prompt selection methods. To obtain the mean IoU (mIoU) we use the same segmentation pipeline as described in \Cref{sec:segmentation_method} changing only the prompt points. The mIoU value is calculated over all 783 grains across the five images. Iterative point selection is able to achieve a higher mIoU in less or the same number of prompt points.

In \Cref{sec:limitations}, we also show that an additional 1-13\% of grains can be found by prompts other than single foreground points. Methods based on optimizing prompt embeddings \cite{zhang2023personalize,Qiu2023} or incorporating semantic information into prompts \cite{Li2023,Wang2023} via text have potential for recovering those grains.

\begin{table*}[t]
  \centering
  \begin{tabularx}{\textwidth}{p{4cm} p{12cm}}
    \toprule
    Characteristic & Description \\
    \midrule
    Space-filling & The entire image is occupied by grains or grain boundaries.\\
    Valley boundaries & Boundary curves are local maxima or minima. \\ 
    Grain compactness & Grains tend to be round.\\
    Boundary Thinness & Boundaries have a maximum thickness. \\
    Boundary Consistency & Boundaries have minimal variation in thickness within a wafer. \\
    Non-overlapping & Grains rarely touch. \\
    Long Tailed & We expect the distribution of grain size to be long-tailed. \\
    \bottomrule
  \end{tabularx}
  \caption{Grain structure characteristics assumed to be inherent to our images}
  \label{tab:characteristics}
\end{table*}

\subsection{Post-Processing}

\begin{table}
  \centering
  \begin{tabular}{m{4.5cm} m{2.5cm}}
    \toprule
    Method & mIoU\\
    \midrule
     SAM's predicted IoU & 0.6195 \\
     Edge Alignment & 0.6309  \\
    \midrule
    difference & + 0.0114\\
    \hspace{1em} Bootstrapped 95\% CI & (0.0021, 0.0207) \\
    \bottomrule
  \end{tabular}
  \caption{Comparison of mask scoring methods for overlap resolution via Non-Maximum Supression}
  \label{tab:sato}
  \vspace{-0.5cm}
\end{table}

Obtaining a good grain mask does not guarantee it will not be filtered in a later pipeline step. For our images, we use Sato edge detection \cite{Sato1998} based on the ``valley boundaries'' characteristic from \Cref{tab:characteristics} to obtain a boolean mask of noisy boundaries. We then derive a goodness score, Edge Alignment, which is the overlap coefficient between the mask perimeter and the noisy boundary. To compare Edge Alignment with SAM's predicted IoU score, we use SAM to generate a set of 50 masks per grain each based on a random foreground prompt point within the grain. For each set of masks we perform NMS to obtain the expected IoU per grain. These results can be seen in \Cref{tab:sato}. The mIoU values are higher than in the previous section due to not competing with masks from prompt points outside each grain. Edge Alignment performs similarly to predicted IoU as a NMS criteria despite not being correlated. However, it does not lead to a large improvement in mIoU.

We have anecdotally seen cases where cropping masks and filling holes can result in good masks not predictable by SAM from single foreground point prompts. This is likely due to the asymmetric boundary detection discussed in \Cref{sec:segmentation_method}. A SAM independent NMS criteria, like Edge Alignment, allows those crops and hole fills to be added post model inference but pre-filtering. However, we have not determined a method for efficiently creating these crop or fill masks.

% \subsection{Image Pre-Processing}

% Changes to the input image and input prompt may allow better output quality. We can follow best practices for preprocessing grain images for image segmentation \cite{Iskakov2020}. 

% Other areas are and dynamically zooming images
% \section{Leveraging SAM for Learning with Limited Labeled Data} % alternatively: training-based improvements
\section{Training-Based Improvements}
\label{sec:fine_tune}

There are many works on fine-tuning SAM \cite{Zhang2023,Chen2023,Archit2023}. These are a great option if you have a representative labeled dataset to train on. Unfortunately, in the design space the expectation of novel images means no training dataset can be truly representative. Furthermore, in this setting, hand-labeled data is potentially more valuable for testing than training. This does not wholly rule out fine-tuning on hand-labeled data but does encourage us to explore alternative approaches. Three are discussed here.

First is data augmentation for teaching specific concepts. Two common errors we have seen qualitatively in other datasets are miss-segmentation due to obscuring contamination and lighting effects. Synthetically introducing these concepts to a diverse training dataset allows for fine-tuning the model without overfitting to a specific domain and losing zero-shot performance.

Second is style-transfer for synthetic data. Previous works \cite{Yu2023,Ruehle2021} have shown success using style transfer to do unsupervised generation of segmentation training data in microscopy domains. This would allow for quick adaptation to new variations in imaging conditions and material texture, eliminating the need for a model with good zero-shot performance and allowing for the use of more traditional fine-tuning methods.

Third is using SAM for weakly-supervised pre-training of a smaller segmentation model. In this scheme, the model is first trained on a large set of in-domain images automatically labeled by SAM then fine-tuned on a few hand-labeled grains. Previous works \cite{Patrick2023,Shi2022} have shown less than 100 images are needed in training U-Net \cite{Ronneberger2015} for segmentation of grains. Domain-specific losses such as clDice \cite{Shit2021} further reduce the amount of training data needed.
\section{Conclusion}
\label{sec:conclusion}

This paper presents an evaluation of Segment Anything as a tool for grain characterization in the hard drive design process. We show experimental results over a limited dataset. The Segment Anything Automatic Mask Generator shows promising out-of-the-box performance at property extraction for our images, but has systematic biases regarding important properties that need to be further explored over a more diverse and representative dataset before deploying it at a larger scale. We also introduce a basic method for determining potential for training-free improvements. Improved prompting and domain-specific scoring show promise for improving segmentation performance and mitigating biases of the AMG pipeline. Future work will refine training-free methods for images of a similar quality to our dataset and explore training-based strategies for poorer quality images.

{
    \small
    \bibliographystyle{ieeenat_fullname}
    \bibliography{main}

\begin{thebibliography}{23}
\providecommand{\natexlab}[1]{#1}
\providecommand{\url}[1]{\texttt{#1}}
\expandafter\ifx\csname urlstyle\endcsname\relax
  \providecommand{\doi}[1]{doi: #1}\else
  \providecommand{\doi}{doi: \begingroup \urlstyle{rm}\Url}\fi

\bibitem[Archit et~al.(2023)Archit, Nair, Khalid, Hilt, Rajashekar, Freitag,
  Gupta, Dengel, Ahmed, and Pape]{Archit2023}
Anwai Archit, Sushmita Nair, Nabeel Khalid, Paul Hilt, Vikas Rajashekar, Marei
  Freitag, Sagnik Gupta, Andreas Dengel, Sheraz Ahmed, and Constantin Pape.
\newblock Segment anything for microscopy, 2023.

\bibitem[Chauveau and Merville(2023)]{Chauveau2023}
Bertrand Chauveau and Pierre Merville.
\newblock Segment anything by meta as a foundation model for image
  segmentation: a new era for histopathological images.
\newblock \emph{Pathology}, 55\penalty0 (7):\penalty0 1017--1020, 2023.

\bibitem[Chen et~al.(2023)Chen, Zhu, Ding, Cao, Wang, Zhang, Li, Sun, Zang, and
  Mao]{Chen2023}
Tianrun Chen, Lanyun Zhu, Chaotao Ding, Runlong Cao, Yan Wang, Shangzhan Zhang,
  Zejian Li, Lingyun Sun, Ying Zang, and Papa Mao.
\newblock Sam-adapter: Adapting segment anything in underperformed scenes.
\newblock In \emph{2023 IEEE/CVF International Conference on Computer Vision
  Workshops (ICCVW)}. IEEE, 2023.

\bibitem[Dillon et~al.(2016)Dillon, Tai, and Chen]{Dillon2016}
Shen~J. Dillon, Kaiping Tai, and Song Chen.
\newblock The importance of grain boundary complexions in affecting physical
  properties of polycrystals.
\newblock \emph{Current Opinion in Solid State and Materials Science},
  20\penalty0 (5):\penalty0 324--335, 2016.

\bibitem[Dosovitskiy et~al.(2021)Dosovitskiy, Beyer, Kolesnikov, Weissenborn,
  Zhai, Unterthiner, Dehghani, Minderer, Heigold, Gelly, Uszkoreit, and
  Houlsby]{dosovitskiy2021an}
Alexey Dosovitskiy, Lucas Beyer, Alexander Kolesnikov, Dirk Weissenborn,
  Xiaohua Zhai, Thomas Unterthiner, Mostafa Dehghani, Matthias Minderer, Georg
  Heigold, Sylvain Gelly, Jakob Uszkoreit, and Neil Houlsby.
\newblock An image is worth 16x16 words: Transformers for image recognition at
  scale.
\newblock In \emph{International Conference on Learning Representations}, 2021.

\bibitem[Herbig et~al.(2015)Herbig, Choi, and Raabe]{Herbig2015}
M. Herbig, P. Choi, and D. Raabe.
\newblock Combining structural and chemical information at the nanometer scale
  by correlative transmission electron microscopy and atom probe tomography.
\newblock \emph{Ultramicroscopy}, 153:\penalty0 32--39, 2015.

\bibitem[Kautzky and Blaber(2018)]{Kautzky2018}
Michael~C. Kautzky and Martin~G. Blaber.
\newblock Materials for heat-assisted magnetic recording heads.
\newblock \emph{MRS Bulletin}, 43\penalty0 (2):\penalty0 100--105, 2018.

\bibitem[Kirillov et~al.(2023)Kirillov, Mintun, Ravi, Mao, Rolland, Gustafson,
  Xiao, Whitehead, Berg, Lo, Dollár, and Girshick]{Kirillov2023}
Alexander Kirillov, Eric Mintun, Nikhila Ravi, Hanzi Mao, Chloe Rolland, Laura
  Gustafson, Tete Xiao, Spencer Whitehead, Alexander~C. Berg, Wan-Yen Lo, Piotr
  Dollár, and Ross Girshick.
\newblock Segment anything, 2023.

\bibitem[Larsen et~al.(2023)Larsen, Villadsen, Mathiesen, Jensen, and
  Boejesen]{Larsen2023}
Rasmus Larsen, Torben~L. Villadsen, Jette~K. Mathiesen, Kirsten M.~Ø. Jensen,
  and Espen~D. Boejesen.
\newblock Np-sam: Implementing the segment anything model for easy nanoparticle
  segmentation in electron microscopy images, 2023.

\bibitem[Li et~al.(2023)Li, Zhang, Sun, Zou, Liu, Yang, Li, Zhang, and
  Gao]{Li2023}
Feng Li, Hao Zhang, Peize Sun, Xueyan Zou, Shilong Liu, Jianwei Yang, Chunyuan
  Li, Lei Zhang, and Jianfeng Gao.
\newblock Semantic-sam: Segment and recognize anything at any granularity,
  2023.

\bibitem[McIntosh(1999)]{McIntosh1999}
John McIntosh.
\newblock Using cd-sem metrology in the manufacture of semiconductors.
\newblock \emph{JOM}, 51\penalty0 (3):\penalty0 38--39, 1999.

\bibitem[Patrick et~al.(2023)Patrick, Eckstein, Lopez, Toderas, Asher, Whang,
  Levine, Rickman, and Barmak]{Patrick2023}
Matthew~J Patrick, James~K Eckstein, Javier~R Lopez, Silvia Toderas, Sarah~A
  Asher, Sylvia~I Whang, Stacey Levine, Jeffrey~M Rickman, and Katayun Barmak.
\newblock Automated grain boundary detection for bright-field transmission
  electron microscopy images via u-net.
\newblock \emph{Microscopy and Microanalysis}, 29\penalty0 (6):\penalty0
  1968--1979, 2023.

\bibitem[Qiu et~al.(2023)Qiu, Hu, Li, and Liu]{Qiu2023}
Zhongxi Qiu, Yan Hu, Heng Li, and Jiang Liu.
\newblock Learnable ophthalmology sam, 2023.

\bibitem[Reinsel et~al.(2017)Reinsel, Gantz, and Rydning]{Reinsel2017}
David Reinsel, John Gantz, and John Rydning.
\newblock Data age 2025.
\newblock Technical report, IDC, 2017.

\bibitem[Ronneberger et~al.(2015)Ronneberger, Fischer, and
  Brox]{Ronneberger2015}
Olaf Ronneberger, Philipp Fischer, and Thomas Brox.
\newblock \emph{U-Net: Convolutional Networks for Biomedical Image
  Segmentation}, pages 234--241.
\newblock Springer International Publishing, 2015.

\bibitem[Rühle et~al.(2021)Rühle, Krumrey, and Hodoroaba]{Ruehle2021}
Bastian Rühle, Julian~Frederic Krumrey, and Vasile-Dan Hodoroaba.
\newblock Workflow towards automated segmentation of agglomerated,
  non-spherical particles from electron microscopy images using artificial
  neural networks.
\newblock \emph{Scientific Reports}, 11\penalty0 (1), 2021.

\bibitem[Sato et~al.(1998)Sato, Nakajima, Shiraga, Atsumi, Yoshida, Koller,
  Gerig, and Kikinis]{Sato1998}
Yoshinobu Sato, Shin Nakajima, Nobuyuki Shiraga, Hideki Atsumi, Shigeyuki
  Yoshida, Thomas Koller, Guido Gerig, and Ron Kikinis.
\newblock Three-dimensional multi-scale line filter for segmentation and
  visualization of curvilinear structures in medical images.
\newblock \emph{Medical Image Analysis}, 2\penalty0 (2):\penalty0 143--168,
  1998.

\bibitem[Shi et~al.(2022)Shi, Duan, Yang, Feng, Ding, and Jiang]{Shi2022}
Peng Shi, Mengmeng Duan, Lifang Yang, Wei Feng, Lianhong Ding, and Liwu Jiang.
\newblock An improved u-net image segmentation method and its application for
  metallic grain size statistics.
\newblock \emph{Materials}, 15\penalty0 (13):\penalty0 4417, 2022.

\bibitem[Shit et~al.(2021)Shit, Paetzold, Sekuboyina, Ezhov, Unger, Zhylka,
  Pluim, Bauer, and Menze]{Shit2021}
Suprosanna Shit, Johannes~C. Paetzold, Anjany Sekuboyina, Ivan Ezhov, Alexander
  Unger, Andrey Zhylka, Josien P.~W. Pluim, Ulrich Bauer, and Bjoern~H. Menze.
\newblock cldice - a novel topology-preserving loss function for tubular
  structure segmentation.
\newblock In \emph{2021 IEEE/CVF Conference on Computer Vision and Pattern
  Recognition (CVPR)}. IEEE, 2021.

\bibitem[Wang et~al.(2023)Wang, Vasu, Faghri, Vemulapalli, Farajtabar, Mehta,
  Rastegari, Tuzel, and Pouransari]{Wang2023}
Haoxiang Wang, Pavan Kumar~Anasosalu Vasu, Fartash Faghri, Raviteja
  Vemulapalli, Mehrdad Farajtabar, Sachin Mehta, Mohammad Rastegari, Oncel
  Tuzel, and Hadi Pouransari.
\newblock Sam-clip: Merging vision foundation models towards semantic and
  spatial understanding, 2023.

\bibitem[Yu et~al.(2023)Yu, Li, Lou, Liu, Wan, Chen, and Li]{Yu2023}
Xinyi Yu, Guanbin Li, Wei Lou, Siqi Liu, Xiang Wan, Yan Chen, and Haofeng Li.
\newblock \emph{Diffusion-Based Data Augmentation for Nuclei Image
  Segmentation}, pages 592--602.
\newblock Springer Nature Switzerland, 2023.

\bibitem[Zhang and Liu(2023)]{Zhang2023}
Kaidong Zhang and Dong Liu.
\newblock Customized segment anything model for medical image segmentation,
  2023.

\bibitem[Zhang et~al.(2023)Zhang, Jiang, Guo, Yan, Pan, Dong, Gao, and
  Li]{zhang2023personalize}
Renrui Zhang, Zhengkai Jiang, Ziyu Guo, Shilin Yan, Junting Pan, Hao Dong, Peng
  Gao, and Hongsheng Li.
\newblock Personalize segment anything model with one shot.
\newblock \emph{arXiv preprint arXiv:2305.03048}, 2023.

\end{thebibliography}
}


\end{document}